# Generalizable semi-supervised learning method to estimate mass from sparsely annotated images

Muhammad K.A. Hamdan[a]  · Diane T. Rover[a] · Matthew J. Darr[b] · John Just[b]

[a] Department of Electrical and Computer Engineering
Iowa State University, Ames, Iowa, USA
[b] Department of Agriculture and Biosystems Engineering
Iowa State University, Ames, Iowa, USA

**Abstract**

Mass flow estimation is of great importance to several industries, and it can be quite challenging to obtain accurate estimates due to limitation in expense or general infeasibility. In the context of agricultural applications, yield monitoring is a key component to precision agriculture and mass flow is the critical factor to measure. Measuring mass flow allows for field productivity analysis, cost minimization, and adjustments to machine efficiency. Methods such as volume or force-impact have been used to measure mass flow; however, these methods are limited in application and accuracy. In this work, we use deep learning to develop and test a vision system that can accurately estimate the mass of sugarcane while running in real-time on a sugarcane harvester during operation. The deep learning algorithm that is used to estimate mass flow is trained using very sparsely annotated images (semi-supervised) using only final load weights (aggregated weights over a certain period of time). The deep neural network (DNN) succeeds in capturing the mass of sugarcane accurately and surpasses older volumetric-based methods, despite highly varying lighting and material colors in the images. The deep neural network is initially trained to predict mass on laboratory data (bamboo) and then transfer learning is utilized to apply the same methods to estimate mass of sugarcane. Using a vision system with a relatively lightweight deep neural network we are able to estimate mass of bamboo with an average error of 4.5% and 5.9% for a select season of sugarcane.

Corresponding author
Iowa State University, Ames IA, USA
✉ E-mail: mhamdan@iastate.edu

## 1. Introduction

Machine learning and computer vision systems have been applied in many agricultural applications such as size estimation of citrus fruit (Shin 2012), crop yield prediction and nitrogen status estimation (Chlingaryan, Sukkarieh, and Whelan 2018), and root-soil segmentation (Douarre et al. 2016). Indeed, the use of machine learning in vision systems has shown much better performance than traditional computer vision approaches (Lee 2015) . Machine learning is the workflow in which features are manually selected and therefore a model is trained.  A subset of machine learning described as deep learning offers automatic learning of features. Deep learning is specifically focused on the use of artificial neural network (architectures that comprise multiple layers) and its variants (Goodfellow, Bengio, and Courville 2016).

Intelligent data analysis techniques such as image classification/identification are used in various agricultural applications (Singh et al. 2016). Some of the applications that deep learning models were employed in include plant classification and identification with convolutional neural networks (Yalcin and Razavi 2016), plant disease recognition by leaf classification (Sladojevic et al. 2016), and classification of land cover and crop types (Kussul et al. 2017).

Supervised learning is a learning paradigm that has been most widely used, but this method often requires large amount of labeled training data. The process of labeling data can be expensive, difficult, and time consuming especially when dealing with very large data sets . In many real world scenarios it is common not to have ground truth measurement for every data point in the dataset, which poses a challenge to supervised learning. Unlike supervised learning,  unsupervised learning methods seek to make use of unlabeled data, but extracting meaningful features without guiding the algorithm in some way is an il-posed problem (Locatello et al. 2018). The algorithm is left to solve the problem through exploring data patterns or generating data clusters. In the presence of sufficient annotated data, unsupervised learning cannot achieve the accuracy levels of supervised learning, since there is no guarantee that the patterns the algorithm finds could correlate directly with the features or characteristics of interest in the data (Lison 2015).

The work presented herein represents a middle-ground between supervised and unsupervised learning, where a version of semi-supervised learning is employed to learn from sparsely annotated data. Semi-supervised learning is motivated by the need for an alternative to the expensive, time-consuming, and tedious process of labeling data as well as to make up for missing ground truth measurements from real world applications. Semi-supervised learning was employed in applications such as crowds counting (Change Loy, Gong, and Xiang 2013), weed mapping in sunflower crops (Pérez-Ortiz et al. 2015), monocular depth map prediction (Kuznietsov, Stuckler, and Leibe 2017), and video object detectors (Misra, Shrivastava, and Hebert 2015).

The work presented herein builds on previous research that applied semi-supervised learning on small logs/runs of bamboo to estimate the mass (and mass flow) in real-time via stream of images. Accurate prediction of mass through images enables precision agriculture through optimizing machine productivity and efficiency. Also, it mitigates risk (e.g. by avoiding overfilling trucks) while offering a more accessible low-cost solution since it only requires a camera and inference via a relatively light-weight neural network. The images in Figure 1 are from the stereo camera mounted on the sugarcane elevator and are used in this work for volume-based and vision-based estimation.

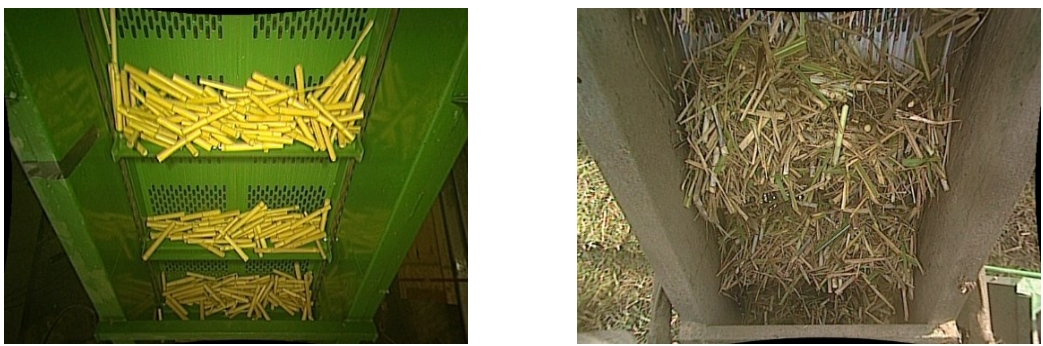

**Figure 1** Left: Example image from lab dataset (bamboo). Right: Example image from field data (sugarcane).

The proposed method can readily be extended to other agricultural applications or industries. Indeed, it is much more widely applicable than what is presented here, but the difficulty of the presented application proves the generality of the approach. In the context of sugarcane mass estimation, the presented deep learning vision based approach is the first ever to be conducted to solve such a problem. Typically, mass is estimated via a direct method such a weight scale (Mailander et al. 2010) or indirectly via estimating volume using optical sensors then converting to mass (Price, Johnson, and Viator 2017). The scale method requires significant changes to the machine, is costly and complex, and susceptible to mechanical noise. Although the optical sensor approach reported strong $R^2$ ranging between 93% and 97%, the system still suffered from the piling up of debris and sugarcane leaves.

Concurrent to the research described herein using deep learning with video data, we investigated an indirect mass estimation method that predicts volume using a stereo camera then converts to mass through a calibrated density. The volumetric method is utilized as a baseline comparison against the vision approach in order to gauge the effectiveness of the proposed vision solution.

This work makes the following **contributions**: 1) Present a robust deep learning approach to learn complex physics relationships between the bulk density, quantity, and location of material in images to accurately estimate mass with only sparse ground truth. 2) Provide a framework that ensures transfer of methods to other material or application with minimal changes. 3) Present a mass estimation method that outperforms volume-based methods, while offering a more cost-effective solution. 4) This work is heavily tested on sugarcane to prove the generalizability of presented methods.

## 2. Materials and Methods

### 2.1 Modified nonlinear regression loss for sparse ground truth

Semi-supervised learning techniques such as self-training, mixture models, co-training, multi-view learning, and 3S-vector machines (Zhu and Goldberg 2009) are commonly used to solve sparsely annotated data problems. Essentially, input data is modeled such that it is utilized to help with the prediction process via (e.g.) data clustering, or self-teaching. In the proposed method, we take advantage of the response itself, where there exist a tractable relationship between each individual data point and the ground truth. In this case, the ground truth represents the total sum of predictions of individual data points for a given period of time.

The typical nonlinear regression formulation using mean squared error (MSE) loss can be slightly modified to incorporate an additional aggregation term over the predictions for each run in "k" runs as shown in Equation (1). The only difference being that the predictions of image "$x_{ij}$" over a given run are summed and then compared to the ground truth and scaled by length "n" of the run. This type of simple modification to nonlinear regression opens up a major opportunity to bypass more costly or infeasible situations to obtain labels or ground truth in order to train a predictive model. The application presented in this work is one such situation where it is much easier to measure an accumulated mass as opposed to trying to obtain an individual mass for each measurement, which would be highly infeasible in any practical sense.

$$L(y;x) = \sum_{i=1}^{k} \frac{1}{n}\left(y_i - \sum_{j=1}^{n} f(x_{ij})\right)^2 \tag{1}$$

### 2.2 Data summary and system setup

#### 2.2.1 Laboratory data

Details of laboratory dataset are referenced for comparison with field data setup. Extensive proof of concept testing was conducted prior to field exposure. Experiments were designed to test the system as closely as possible to factors present during typical operation, while controlling for factors outside of system control such as particle density. To accomplish this, bamboo was used as a surrogate material to sugarcane since it has stable material properties long term (will not rot or dry out) while being similar in shape to sugarcane. Bamboo was conveyed into a sugarcane elevator at various flow rates with a stereo camera mounted on top of the elevator. The testing factors included variable illumination, material flows, and elevator speeds. A total of 239 runs were split into {60, 20, 20} - {train, validation, test} sets, which include 8 empty runs (zero mass) running for ~ 60 seconds with the elevator moving. Figure 2 shows the overall system of laboratory setup, which includes a hydraulically driven elevator and conveyor, a logging system, a stereo camera system, and a scale. The laboratory setup is identical to the field setup except that the ground truth weights (aggregated weights of a complete run) came from the scale under the bamboo hopper instead of a weight wagon and did not involve wireless transceivers.

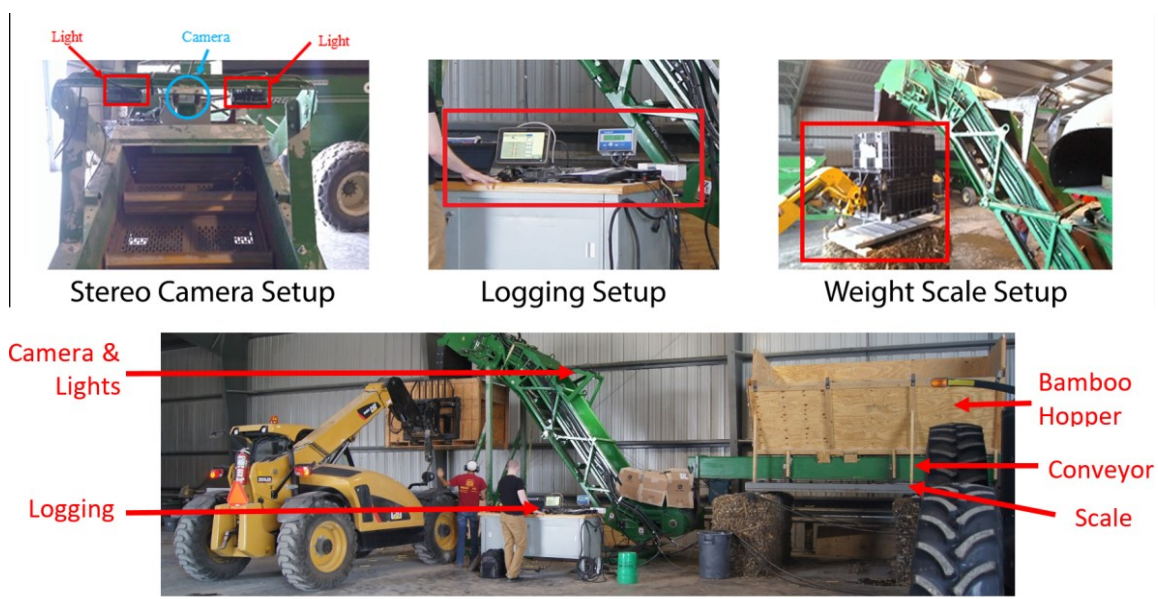

**Figure 2** Overall system showing the components used in laboratory testing

### 2.2.2    Field data

Field data was collected in the course of three consecutive years (2014 through 2016) and in four different regions (Brazil, Florida, Louisiana, and Texas) to test robustness of the system to various environmental factors that could influence bulk and particle densities of the material. Data is comprised of 1567 runs/wagon loads (over 3M images), and contains both burnt and green sugarcane. This was an important part of the design of experiment since burnt cane, in which the leaves and fibrous trash were burnt off, was projected to represent the high end for density, whereas green cane was expected to vary more depending on the amount of trash and ability of the primary extractor fan to remove trash. Table 1 summarizes runs distribution based on location, season of harvest, and type of sugarcane.

**Table 1** Runs distribution based on years, region of harvest, and material type

| Crop Type | Region | Sugarcane Harvest Year | | | Total Region |
|---|---|---|---|---|---|
| | | 2014 | 2015 | 2016 | |
| Green | Louisiana | | 669 | | 669 |
| | Brazil | 166 | | | 166 |
| | Florida | | | 264 | 264 |
| Burnt | Texas | | 145 | | 145 |
| | Florida | | | 323 | 323 |
| Total | | 166 | 814 | 587 | **1567** |

Wireless transceivers were used to collect ground truth from scales, which were installed on the wagons as shown in the diagram in Figure 3. The wagons were typically six or nine metric ton capacities. An image processing unit (stereo camera + algorithm) was used to generate colored images and 3d point cloud (converted into volume) of material. A speed sensor was used to capture the elevator speed. The image processing unit recorded the speed sensor, machine states, and operating points via CAN bus. Image data was transferred to a dedicated stereo logger via an Ethernet link between the image processing unit and the stereo logger.

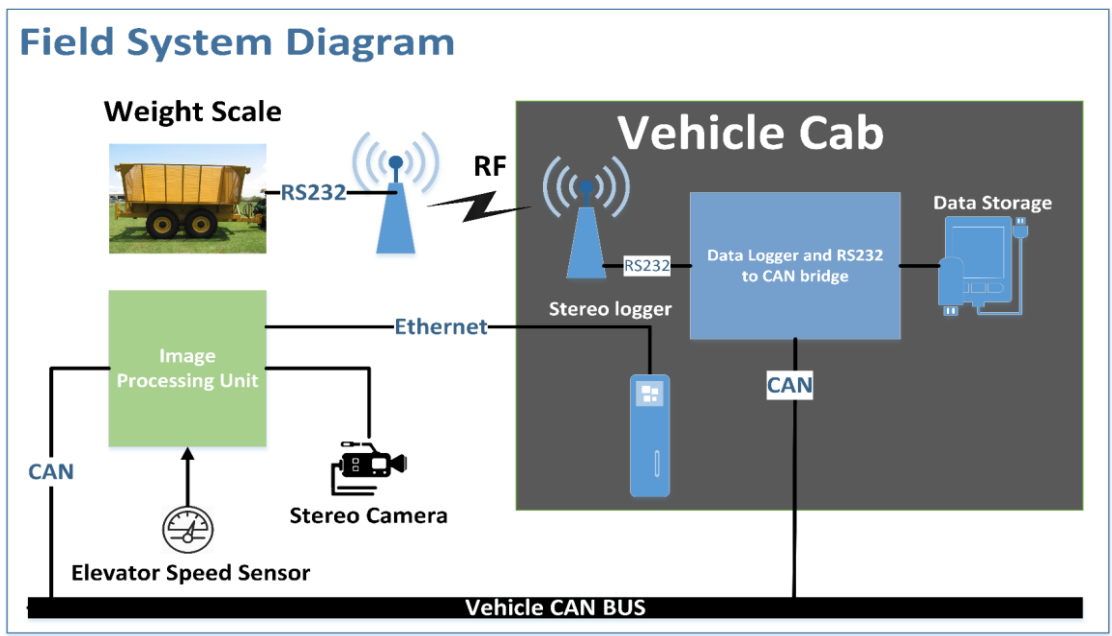

**Figure 3** Block diagram of the field system setup highlighting the components used for data generation and logging.

### 2.3　Data complexity

Training a deep learning algorithm to estimate mass flow from video data is far more complex for the in-field harvesting application than a controlled laboratory scenario. Laboratory data or bamboo has a consistent yellowish color and the elevator background is always green, yet the only factor that affects bamboo images is lighting. Figure 4 shows images of bamboo under different ambient lighting conditions.

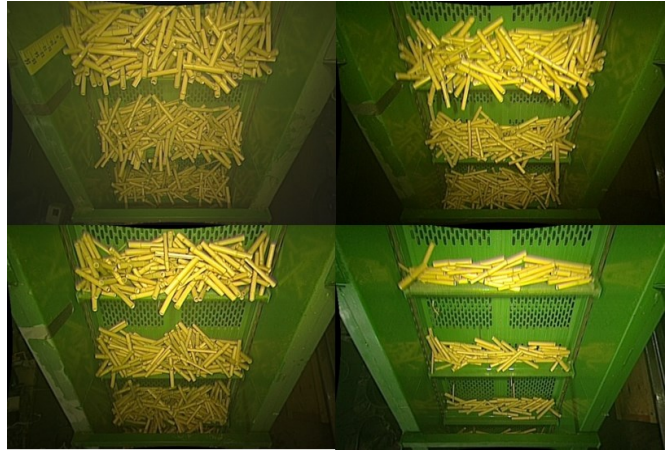

**Figure 4** Bamboo images under different ambient light conditions

Unlike laboratory experiments, field experiments were affected by more extreme lighting variation, shadowing, dirt (on lens, material, background), elevator background color, material composition (e.g. green, burnt, root balls), billet and leaf size and color variation, larger changes in material density, extractor fan presence, airborne debris, and grossly overfilled slats. Figure 5 shows a set of images of the extractor fan blocking the view of the camera. It can be seen that images also have different colors and lighting conditions.

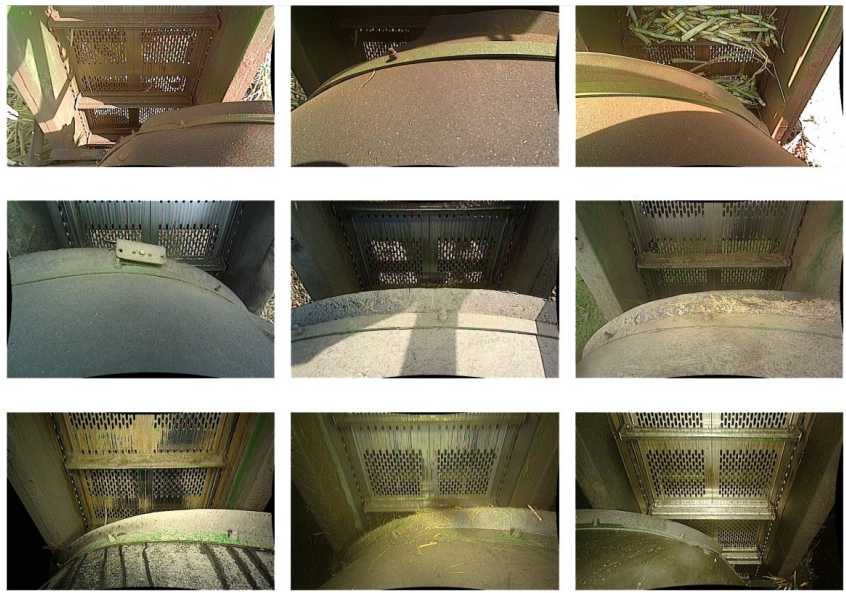

**Figure 5** Different images of the extractor fan blocking the view of the camera.

Even without the presence of the extractor fan the elevator running empty takes on many different colors due to dirt, rust, stripped paint, and exposure to varying lighting source as shown in Figure 6. Lastly, a set of images of different sugarcane content are shown in Figure 7. Observing these images it is shown how much more complex the sugarcane application is compared to the proof of concept (bamboo) testing in Figure 4. Regardless, in this work it is demonstrated that the devised algorithm based on laboratory data can still learn to estimate mass despite all of these different confounding factors. Another complicating factor to consider is the length of field logs/runs, which can be up to 26x longer than laboratory logs/runs. Thus, this method must learn to predict mass from very sparse ground truth, which affects the ability of convergence during training.

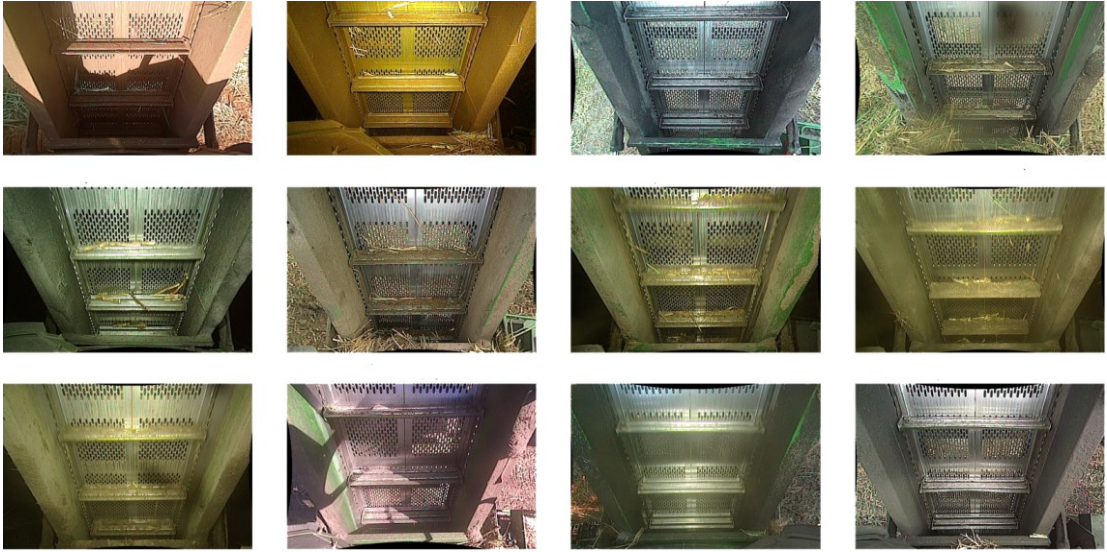

**Figure 6** Empty elevator images with various colors and under different lighting conditions.

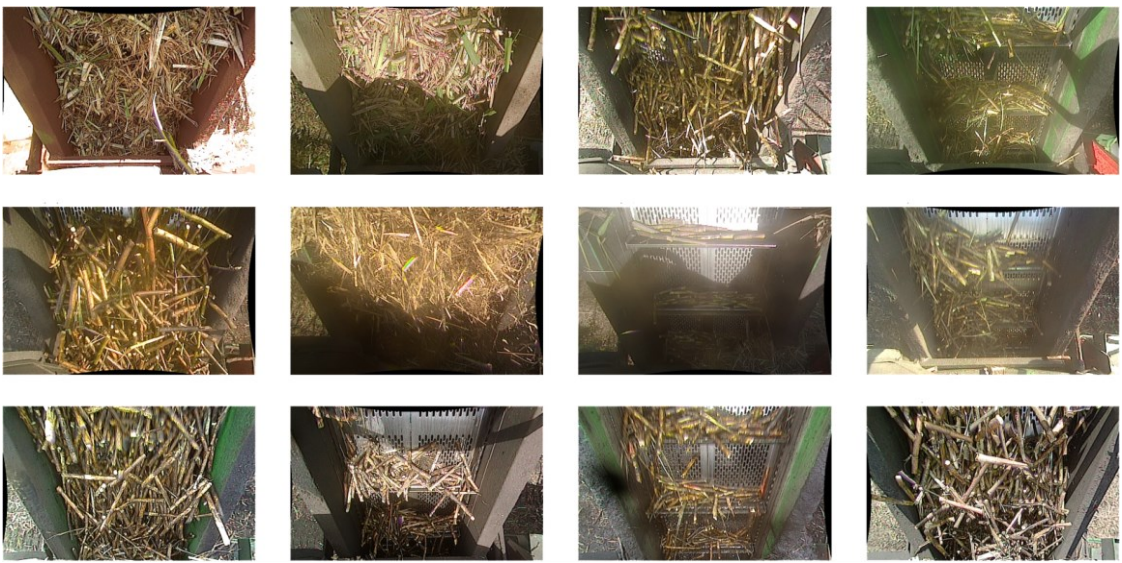

**Figure 7** Different sugarcane content (green and burnt) with various amounts shown to have different colors as exposed to sunlight.

## 2.4　Mass estimation algorithms

### 2.4.1　Mass estimation by volume

A volume estimate obtained from a stereo camera rated at 7.5Hz is used as a baseline comparison for our implementation. By producing point cloud referenced to the plane of the elevator from the stereo camera, volume was calculated by binning the point cloud into squares in the $<x, y>$ plane. To remain insensitive to outliers, the median per bin was considered. The stereo camera estimates the volume within the region of interest (ROI), which is denoted as "$V_c$", then this quantity is scaled by the distance the elevator moves ($\Delta t \times V_e$) in between the next volume estimate to find an incremental accumulated volume. A simplifying assumption inherent in this formulation is that the volume calculated "$V_c$" is spread evenly across the ROI, since the incremental accumulated volume "$V_\Delta$" is directly proportional to the distance the elevator moves. The incremental volume is then converted to mass via a multiplier (density) as seen in Equation (2). Any error in density ($\rho$) of the material can be seen to directly contribute to error in predictions of mass ($m_\Delta$).

$$V_\Delta = \Delta t \times V_e \times V_c \; ; \; m_\Delta = V_\Delta \times \rho \tag{2}$$

To convert volume measurements to mass, an algorithm was devised and fit that explicitly uses the incremental volume to predict the density, and then use it as a multiplier on the volume to convert to mass as shown in Equation (3).

$$Mass = f(\max(V - \beta, 0); \theta) \times \max(V - \beta, 0) \times V_e \times \Delta t \tag{3}$$

Where "$f$" is a feed-forward neural network parameterized by "$\theta$" that outputs a prediction of density based on the raw volume "V", scaled by elevator speed "$V_e$" and capture time "$\Delta t$". The neural network is composed of 4-layers with a total of 256-hidden units (32-64-128-32). The $\beta$-parameter is also fit to account for any positive bias present since the stereo volume calculation was designed towards not missing any volume. This was also a meaningful formulation since if the "$\beta$" parameter turned out to be negative then it would indicate a volume estimation is in need of refinement, since it would be compensating for volume not detected.

### 2.4.2 Learning mass from images

Learning mass from images is possible in this problem context due to the constraints imposed on the system and the ability to learn complex patterns via end-to-end training with deep learning. More specifically, the camera has a fixed sampling frequency that is fast enough to measure all the material passing by. Further, the camera is mounted at a fixed distance from the elevator, and the only additional factor needed is slat velocity to scale the accumulated mass. Intuitively, the velocity is scaling the mass to produce a mass flow, since if the mass was sitting still it would not be accumulated. Alternatively, it can be thought of it as a way to account for frame overlap.

Even though complex nonlinearities exist with the density and the material location in the image (e.g. the same material further away is smaller in the image), these patterns can be learned by optimizing a DNN due to fixed locations between the camera and the elevator. Understanding this reasoning about the system is crucial to designing a successful model in this and similar applications. Assuming counting pixels of material will correlate well with mass would lead to wasted efforts considering only a factor such as image deformation.

The formulation to predict mass from images is shown in Equation (4), where "$f$" is a deep residual convolutional neural network that is parameterized by "$w$". "$x_{ij}$" represents input image "$j$" in $run_i$ and "$V_{ij}$" is elevator speed at time "$j$" in $run_i$ and "$\Delta t$" is the time per frame.

$$Mass_{ij} = f(x_{ij}; w) \times V_{ij} \times \Delta t \tag{4}$$

To be able to predict mass using the deep convolutional neural network the modified loss function introduced in Equation (1) is used and only the scaling factor ($V_{ij} \times \Delta t$) is added to it as shown in Equation (5).

$$L(y; x; w) = \frac{1}{n_i}\left(y_i - \sum_{j=1}^{n} f(x_{ij}) \times V_{ij} \times \Delta t\right)^2 \tag{5}$$

### 2.4.2.1 Temporal smoothing

Utilizing prior knowledge about the problem allows more custom formulation of the model, loss function, and training procedure to improve accuracy and stability. In this application, it is clear that there is temporal correlation. Images near in time should have more similarity in mass than images further away in time. To account for this during training, a regularizing term was added to the cost function with an associated hyper-parameter that allows penalizing a 1st order lagged difference in the predicted mass values. Equation (6) shows full loss function for a $run_i$ that includes prediction error and the additional temporal smoothing regularization term in red.

$$L(y; x; w) = \frac{1}{n_i}\left(y_i - \sum_{j=1}^{n} f(x_{ij}) \times V_{ij} \times \Delta t\right)^2 \\ + \frac{\lambda}{n}\sum_{j=1}^{n_i}\{f(x_{ij}; w) - f(x_{i(j-1)}; w)\}^2 \tag{6}$$

Prediction is corrected by the elevator speed "$V_{ij}$" and capture time "$\Delta t$" (constant 7.5Hz) to account for frame overlap. The loss is normalized by the number of images "$n$" in each run to equally weight the gradient update from each run. To see this, note that the gradient of the loss function has a sum of gradients in it, if the gradient is left un-normalized, it will amount to larger gradient updates for longer runs (runs with more images) even if runs contain the same total mass. The penalty strength is controlled by a hyper-parameter $\lambda$ (chosen experimentally - 0.05).

### 2.4.2.2   Gradient update

To perform a gradient update, predicted value "$\hat{y}_i$" needs to be compared with ground truth "$y_i$", but prediction and regularization terms contain sums, which means the gradient of the loss does as well, as can be seen by Equation (6) . Since the number of images is too large to fit on a single GPU (a problem for very sparse ground truth), a running sum of the gradients, predictions, and regularizing term is maintained from each batch to greatly reduce memory requirements. Equation (7) describes gradient of the full form of the loss equation associated with $run_i$. Terms in red are accumulated from each batch and the full gradient is calculated and applied when the end of a run is reached during the training loop.

$$\frac{\partial L_i}{\partial w} \leftarrow \frac{2}{n_i}\left[y_i - \sum_{j=1}^{n_i} \hat{y}_{ij}\right] \times \sum_{j=1}^{n_i} \frac{\partial \hat{y}_{ij}}{\partial w}$$
$$+ \frac{2\lambda}{n_i} \sum_{j=1}^{n_i}\left\{\left[\breve{y}_{ij} - \breve{y}_{i(j-1)}\right] \times \left[\frac{\partial \breve{y}_{ij}}{\partial w} - \frac{\partial \breve{y}_{i(j-1)}}{\partial w}\right]\right\}^2 \quad (7)$$

Given the large and variable size of runs, gradients and predictions are computed in batches and accumulated over the course of their respective runs as shown in the Pseudo-code in Algorithm 1.

---

**Algorithm 1** Computing and applying gradients over a single epoch. For each $run_i$, run length, images and elevator speeds, and total ground truth mass are available.

```
1:   for run in train_data do:
2:     for batch, (images, speeds) in run do:
3:       x_b, v_b = fetch_next_batch(images, speeds)        ▷ Data loaded in chronological order
4:       If New_run then:
5:         Run_remainder ← mod(run_length, sizeof(batch))
6:         Iterations ← ceil(div(run_length, sizeof(batch)))
7:         New_run ← False
8:       If iterations > 1 then:
9:         ŷ_b ← f(x_b; w)                                  ▷ Predict using DNN
10:        ŷ_{b_vt} ← ŷ_b × v_b × t                          ▷ Correct frame overlap
11:        ŷ += Σŷ_{b_vt}                                    ▷ Accumulate predictions
12:        ŷ_{b_grad} += Σ ∂ŷ_{b_vt}/∂w                      ▷ Accumulate gradients
13:        ŷ_smooth += Σ{(ŷ_{b_vt} - ŷ_{b_vt-1}) ⊙ (∂ŷ_{b_vt}/∂w - ∂ŷ_{b_vt-1}/∂w)}   ▷ Accumulate penalty
14:        Iterations -= 1
15:      Else:
16:        If batch contains current_run_images_only then:
17:          ∂L_i/∂w ← -(2/n_i)(y_true - ŷ) ⊙ ŷ_{b_grad} ⊕ ŷ_smooth                  ▷ Apply gradients
18:          w ← w + α ∂L_i/∂w                                                        ▷ Update weights
19:          New_run ← True
20:        Else:
21:          Do steps 17 → 19 for images in current_run
22:          Do steps 3 → 14 for images in next_run
```

## 3. Results and Analysis

### 3.1  Model architecture and training procedure

Complexity and size are considered as essential factors in the developed architecture, where eventual application goes on embedded hardware at mass scale and saving every bit of computation to a minimum is highly desirable to save costs. A residual like deep neural network is adopted and a systematic approach in DNN design is followed, starting shallow and then reproducing accordingly. A 9-layer DNN with ELU units and residual connections, defined as – "RES9-ER" shown in Figure 8, was found to have good predictive accuracy as well as fast training and inference on laboratory data. Similar models with ReLU activation and 16 layers were also investigated before converging on this final architecture. ELU activation function was considered in the DNN because it showed better noise dampening and more stable signal as well as helped converge faster than ReLU activation.

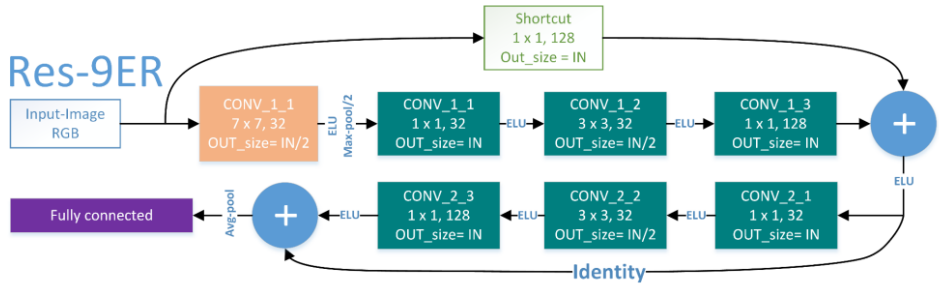

**Figure 8** Reduced residual 9 architecture (RES-9ER).

The 16-layer network - "RES-16E", barely out-performed the 9-layer network - "RES-9E". Thus, the learned features were investigated in every layer of "RES-9E" and redundant features were observed. Therefore, the number of filters in "RES-9E" were reduced to introduce the best performing architecture defined as - "RES-9ER". "RES-16E" has a total of 959,489 parameters and RES-9E has 154,113 parameters, while "RES-9ER" has only 45,921 parameters. "RES-9ER" inference time for a batch size of 8 is ~348FPS when running on 1080-Ti GPU, and ~91FPS when running on Intel Core-i7-7600U CPU. The FPS is proportional to batch size (i.e. sampling frequency) of images and subject to I/O or memory bounds.

Field data was trained and tested on each individual region per Table 1, and then on the data as whole (a more desirable scenario). Transfer learning was utilized with the vision-based approach to cut the training times since training using a deep residual neural network with images takes fairly a long time (around 2 days in the case of All-fields combined) given the large dataset (over 3M images). To optimize training speed, Tensorflow DATASET API was used to create the input pipeline and data was preprocessed and stored in the format of Tensorflow TFrecords.

Initially, field data was trained using the "RES9-ER" model, but opportunities such as adding dropout layers to the model were explored to see if it would help the model generalize better; however, adding dropout did not seem to have any effect on the model performance and therefore was discarded. Furthermore, training on gray-scale images was considered, which did not obtain the level of accuracy as when the model was trained on colored images. Moreover, when attempted to train on gray-scale images and the penalty term (temporal smoothness) was removed, that resulted in noisy signals as the example shown in Figure 9.

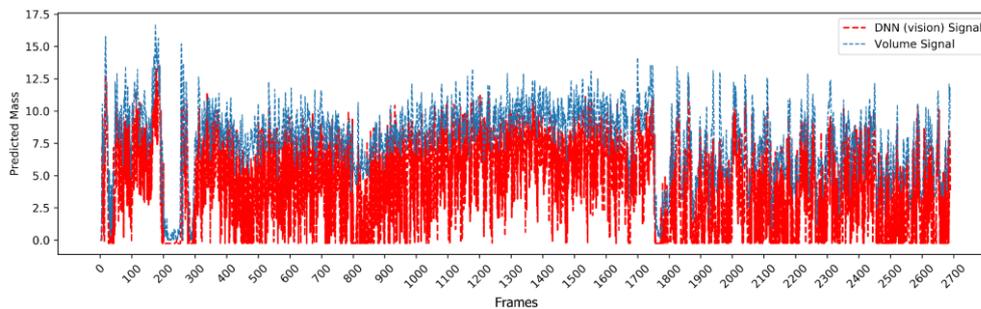

**Figure 9** Overlay of vision estimated signal over volume estimated reference signal of a select run/log. The vision signal was predicted using gray-scale images and no temporal smoothing was applied to the signal

Training on "RES-9E" was also considered but it did not seem to perform any better than "RES-9ER". Further, batch-normalization layers (Ioffe and Szegedy 2015) were used with deeper architectures; however, they enforced an undesirable smoothing effect (shown in Figure 10) that affected the overall shape of the prediction signal, hence impacting the accuracy. In conclusion and after numerous training trials, the best performing architecture remains "RES-9ER" with only the penalty term added.

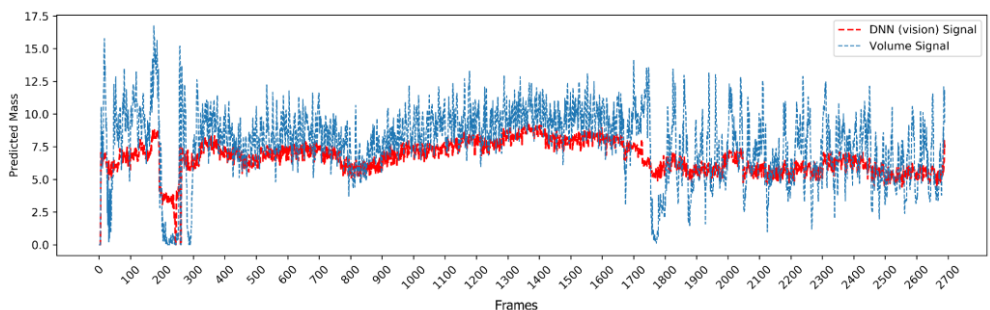

**Figure 10** Overlay of vision estimated signal with volume estimated reference signal of a select run/log. The vision signal was predicted using a 16 layer architecture with batch normalization layers added

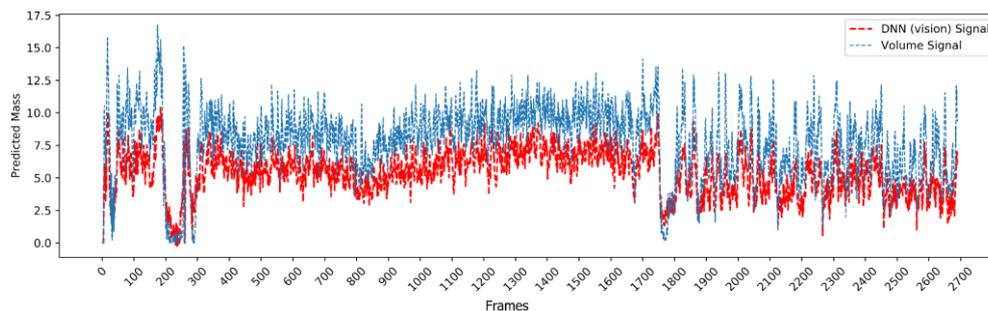

**Figure 11** Overlay of vision estimated signal with volume estimated reference signal of a select run/log. The vision signal was predicted using RES-9ER, and it shows a reasonable shape.

## 3.2　Predicted signals investigation

Scoring accurate predictions in terms of how close the accumulated mass compared to the ground truth measurement of each run/log is feasible and interpretable; however, it does not guarantee correct predictions of mass flow per image since the chance exists that the model could learn non-generalizable patterns. To investigate that the model was learning the proper features and going to generalize, the output signal from the DNN was compared against the volume estimation signal from the stereo camera for several runs. It is emphasized that, while the volume signal may not accurately reflect mass without an accurate density estimate, the relative shape of the volume signal is going to correlate with mass flow strongly. Thus, the volume signal can be used to roughly check that the mass flow estimate from images is reasonable.

Figure 12 shows an example run where the DNN misidentifies empty spots and other regions (circled in green). This run is from Florida dataset, which initially lacked empty or zero runs in it so the network could not learn such features. To remedy this problem, new empty runs were sourced and the DNN was retrained to give a better overall signal shape as shown in Figure 13.

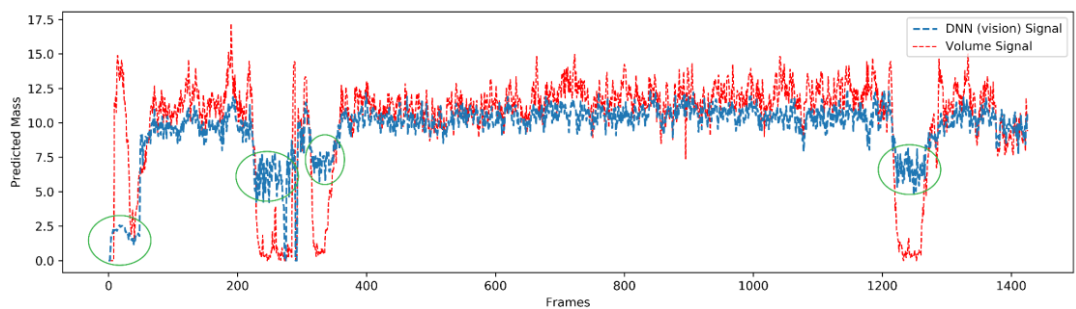

**Figure 12** Training without zero runs resulted in an incorrect overall prediction of approximate signal shape.

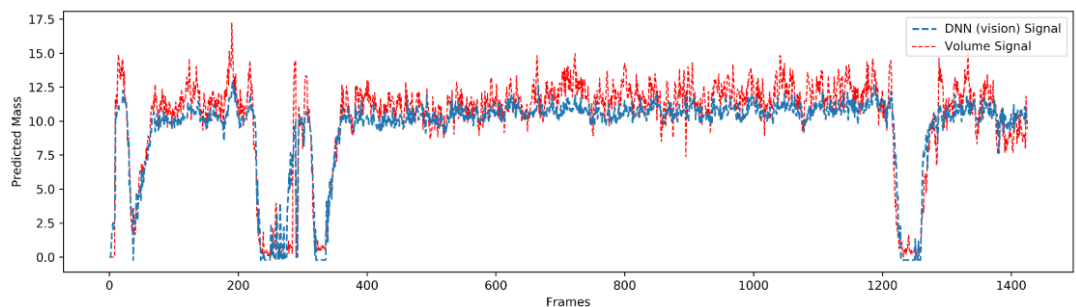

**Figure 13** Better prediction of signal shape is maintained with the use of empty runs.

Another instance that was observed to affect the signal shape is when the extractor fan blocks part of the camera view as shown in Figure 14. This situation occurs when the elevator is running at very low speed or stationary with usually no flowing material.

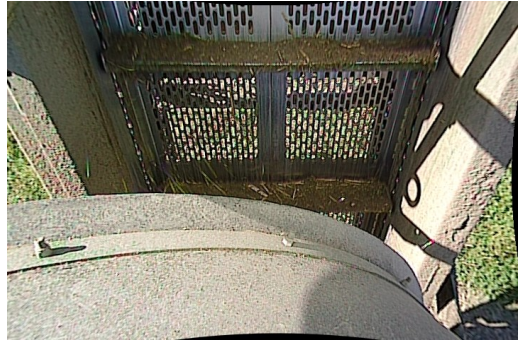

**Figure 14** Image of the extractor fan from an example run/log that is affected from the presence of the extractor fan.

The presence of the extractor fan makes the DNN add noise to the predicted signal (area circled in red in Figure 15) at the empty spots, where it is supposed to be straight flat. This is not an extreme situation and does not significantly impact the accuracy of the system overall, and can be remedied by referencing a signal of the extractor fan and elevator positions via CANBus.

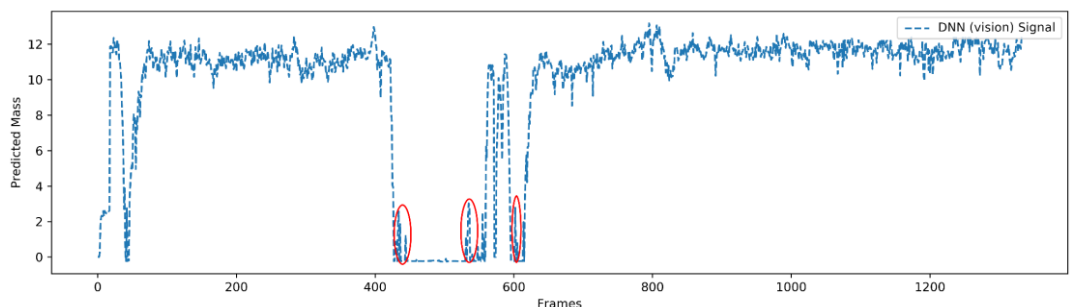

**Figure 15** Noise observed at the empty spots in the predicted signal as the extractor fan blocks part of the camera view.

To further demonstrate that material flow is being captured correctly, Figure 16 through **Error! Reference source not found.** show overlays of the predicted vision signal on top of the reference volume signal for a select run from each region. The selected runs show intermittent and incremental/decremental flows. In all scenarios, it is shown that the vision signal is smoother than the volumetric signal and that is due to applying the temporal smoothing term in the learning process as well as the learned features from images are consistent especially when no material is present. It is seen in Figure 16 that the DNN successfully captures the sudden presence of material (spikes).

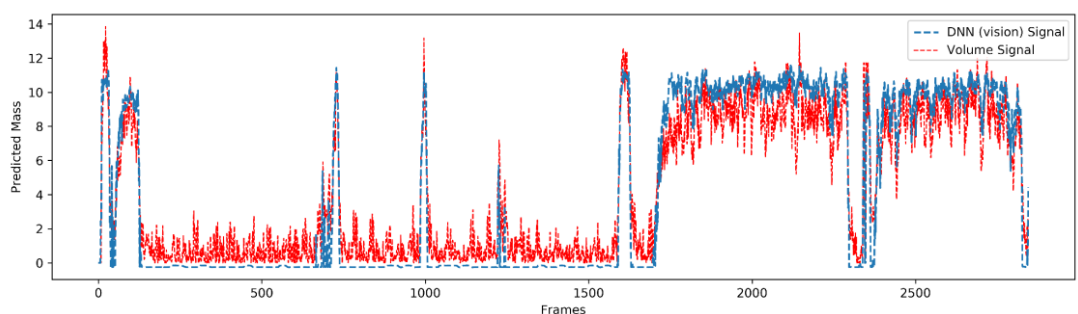

**Figure 16** Selected run from Louisiana dataset showing intermittent spiky material flow.

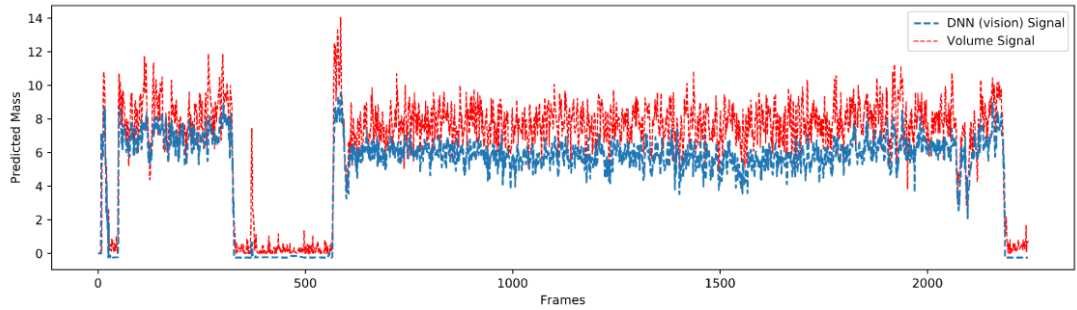
**Figure 17** Selected run from Texas dataset showing consistent material flow.

Figure 18 shows that the DNN signal successfully captures the fine details of the signal shape as material flow changes with time (increase and decrease). The vision and volumetric signals correlate a lot with each other except that vision signal is smoother and that is again due to the temporal smoothing effect.

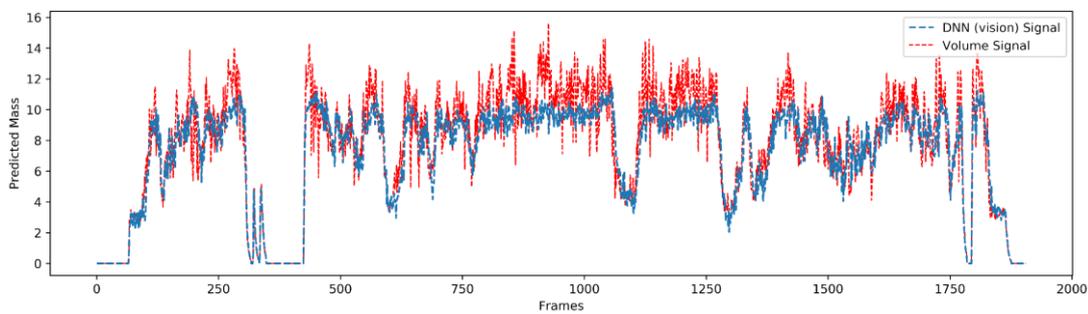
**Figure 18** Selected run from Brazil dataset showing incremental and decremental material flow.

### 3.3 Overall performance

Field data (sugarcane) was evaluated using the same methods used with laboratory data (bamboo). The core training procedure for the field data followed the same procedure as the laboratory data highlighting that the proposed semi-supervised method is transferable to other crops without further changes. Field data that is evaluated using the volumetric-based approach was trained on a 4-layer feed-forward neural network to adjust for nonlinear changes in material density that correlate with volume. Volume estimates serve as a baseline comparison to the vision-based deep learning method.

The vision-based DNN learns mass directly from images and then mass is scaled by elevator speed and capture time to account for frame overlap. Table 2 summarizes the average error of both the volumetric and the vision-based approaches for the different regions independently and for all the data as whole. Note that in order to convert volume to mass, a calibration coefficient (density) is estimated via a feedforward neural network and hence used a multiplier. Some extreme outliers were found the case of Florida and All-fields combined and were excluded from reported average error, but are discussed in the outliers section (Figure 21). As a comparison, these same methods were applied to the bamboo data (laboratory experimentation) achieving an average error of 4.5%.

Table 2 Test-set average error (MAE) per region

| Region | Test-set Average Error | |
|---|---|---|
| | Vision | Volumetric |
| Louisiana | 6.22% | 10.54% |
| Texas | 5.88% | 7.78% |
| Brazil | 8.76% | 8.86% |
| Florida | 12.83% | 14.69% |
| All-Fields | 12.97% | 23.38% |

From Table 2 it can be seen that the vision approach outperforms the volumetric approach in each of the fields and when All-fields are combined. Florida seemed to have high average error compared to other fields because it included both burnt and green cane as well as due to the presence of high trash content. Similarly, the volumetric approach fared poorly when evaluated on All-fields and that is due to the nature of volumetric signal being it highly dependent on the estimated density which varies by material type (green or burnt). Variations between fields exist due to the different machine operation and environmental conditions.

Given the different environmental factors and machine operation settings, the vision approach offered better average error when combining All-fields from different regions together with ~10% improvement compared to the volumetric All-fields average error. This demonstrates the robustness of the vision-based method when posed with different types of material and environmental conditions.

### 3.4  Outliers investigation

The camera system worked well even with very little maintenance, up to and including when the lens on the camera were severely covered with dirt or dust. However there were a few rare instances when a perfect storm occurs and external factors affect the camera performance. These conditions are hard to remedy and affected both volume and vision estimates, but fortunately as mentioned were found to be rare.

An instance was found for a short period of a day when harvester was driven in a particular direction, the sun would shine directly into the camera and wash out portion of the point cloud or add shine blob to images. Another instance that occurred when material flows were very high and the trash, especially dry and light, could fly up and block the camera view as shown in Figure 19.

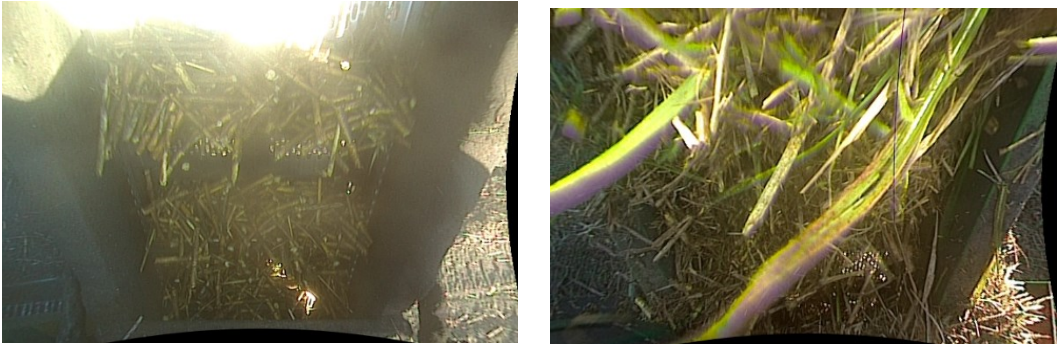

**Figure 19** Left: Sunlight washing out portion of the images. Right: Trash flying up and blocking the view of the camera.

Generally, the vision system estimates were biased low for some runs that had large sunshine blob present in the images, and biased high when there were high trash content in images. The latter situation can be improved in future studies where another algorithm can work concurrently to estimate the amount of trash present and then use such information in the DNN mass estimation algorithm. Figure 20 shows overlay of error distributions of both vision and volume estimates of All-fields test set. The outliers (circled in red) in the distribution were due to bad speed sensor measurements.

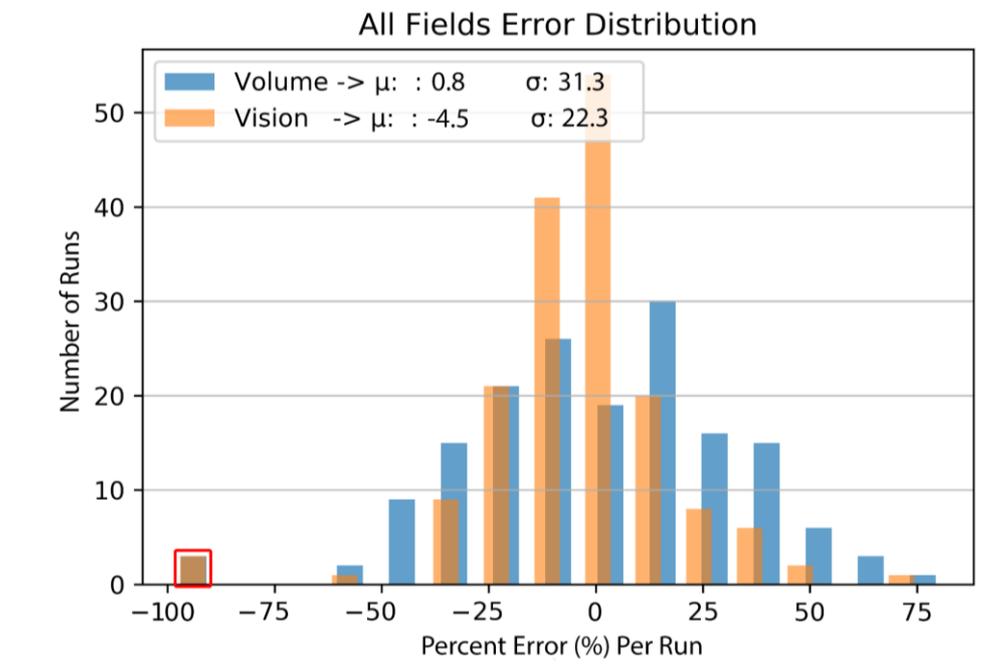

**Figure 20** Histogram distribution of error of volume and vision estimates on all-fields test set.

To further investigate other potential outliers as well as under/over estimates, the error distribution relationship with mass flow was studied for the All-fields test set. As shown in Figure 21, some runs were identified to have bad speed sensor measurement, one run has bad ground truth measurement, one run was under-estimated due to being exposed to direct sunlight, and a few runs were over-estimated due to the presence of high trash content. It is again emphasized that the cases of high trash content can likely be remedied with inclusion of trash prediction algorithm in the training process.

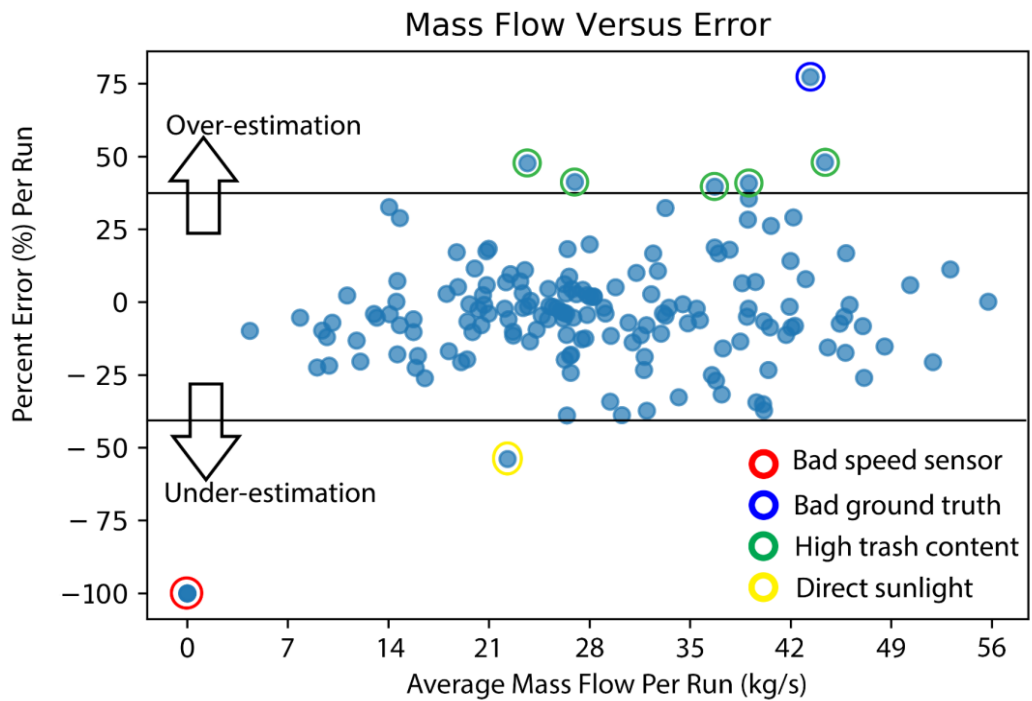

**Figure 21** Mass flow versus error for all-fields test set.

### 3.5 Guidelines to knowledge transfer

To ensure a successful transfer of knowledge, a list of guidelines is set. Following these guidelines shall support the transfer of the methods presented in this work to other crops or perhaps applications. These guidelines were derived from extensive experimentation and research as we progressed in solving the problem at hand. The guideline are summarized as follows:

1) Ensure that there is sufficient number of empty runs as part of the design of experiment. This is critical factor to identifying empty spots in runs.

2) Ensure that the camera is at fixed suitable location from material.

3) It is preferred not to have lengthy runs in the design of experiment. Shorter runs means less sparse ground truth which helps the optimization process learn faster.

4) It is preferred to have consistent run lengths as this would help the optimization process learn faster. What matters is obtaining accurate prediction on the data point level not on the run level.

5) It is recommended to have 60/20/20 - train/test/validation data split for small data sets (like the laboratory dataset) and 80/10/10 for large data sets (like the field dataset).

6) If designing a new DNN architecture, including batch normalization would help the network converge especially if it is deep but it might affect the shape of predicted signal, hence the accuracy.

7) If designing a new DNN architecture, adding dropout layers might help prevent over-fitting but may not improve accuracy.

8) Training with gray-scale images is possible and can score relatively good compared to colored images.

## 4. Conclusion

In this work we presented a generalizable semi-supervised algorithm that makes inference on mass flow of material from sequences of images by training a deep neural network with very sparse ground truth. The presented vision method showed improvements over older and more expensive methods that must first acquire a volumetric estimate of the material and then calibrate using density. The presented algorithm was tested on two different materials and under controlled and real operation environments. The results obtained herein demonstrate that the algorithm is readily transferable to other crops or even applications. Improvements to the system can be further obtained by incorporating a trash estimator in the training process.

# 5. References

Change Loy, Chen, Shaogang Gong, and Tao Xiang. 2013. "From Semi-Supervised to Transfer Counting of Crowds." In *Proceedings of the IEEE International Conference on Computer Vision*, 2256–63.

Chlingaryan, Anna, Salah Sukkarieh, and Brett Whelan. 2018. "Machine Learning Approaches for Crop Yield Prediction and Nitrogen Status Estimation in Precision Agriculture: A Review." *Computers and Electronics in Agriculture* 151: 61–69.

Douarre, Clement, Schielein Richard, Carole Frindel, Stefan Gerth, and David Rousseau. 2016. "Deep Learning Based Root-Soil Segmentation from X-Ray Tomography." *BioRxiv*, 71662.

Goodfellow, Ian, Yoshua Bengio, and Aaron Courville. 2016. *Deep Learning*. MIT press.

Ioffe, Sergey, and Christian Szegedy. 2015. "Batch Normalization: Accelerating Deep Network Training by Reducing Internal Covariate Shift." *ArXiv Preprint ArXiv:1502.03167*.

Kussul, Nataliia, Mykola Lavreniuk, Sergii Skakun, and Andrii Shelestov. 2017. "Deep Learning Classification of Land Cover and Crop Types Using Remote Sensing Data." *IEEE Geoscience and Remote Sensing Letters* 14 (5): 778–82.

Kuznietsov, Yevhen, Jorg Stuckler, and Bastian Leibe. 2017. "Semi-Supervised Deep Learning for Monocular Depth Map Prediction." In *Proceedings of the IEEE Conference on Computer Vision and Pattern Recognition*, 6647–55.

Lee, Andy. 2015. "Comparing Deep Neural Networks and Traditional Vision Algorithms in Mobile Robotics." *Swarthmore University*.

Lison, Pierre. 2015. "An Introduction to Machine Learning." *Language Technology Group (LTG), 1* 35.

Locatello, Francesco, Stefan Bauer, Mario Lucic, Gunnar Rätsch, Sylvain Gelly, Bernhard Schölkopf, and Olivier Bachem. 2018. "Challenging Common Assumptions in the Unsupervised Learning of Disentangled Representations." *ArXiv Preprint ArXiv:1811.12359*.

Mailander, Mike, Caryn Benjamin, Randy Price, and Steven Hall. 2010. "Sugar Cane Yield Monitoring System." *Applied Engineering in Agriculture* 26 (6): 965–69.

Misra, Ishan, Abhinav Shrivastava, and Martial Hebert. 2015. "Watch and Learn: Semi-Supervised Learning for Object Detectors from Video." In *Proceedings of the IEEE Conference on Computer Vision and Pattern Recognition*, 3593–3602.

Pérez-Ortiz, Maria, J M Peña, Pedro Antonio Gutiérrez, Jorge Torres-Sánchez, César Hervás-Mart\'\inez, and Francisca López-Granados. 2015. "A Semi-Supervised System for Weed Mapping in Sunflower Crops Using Unmanned Aerial Vehicles and a Crop Row Detection Method." *Applied Soft Computing* 37: 533–44.

Price, Randy R, Richard M Johnson, and Ryan P Viator. 2017. "An Overhead Optical Yield Monitor for a Sugarcane Harvester Based on Two Optical Distance Sensors Mounted above the Loading Elevator." *Applied Engineering in Agriculture* 33 (5): 687–93.

Singh, Arti, Baskar Ganapathysubramanian, Asheesh Kumar Singh, and Soumik Sarkar. 2016. "Machine Learning for High-Throughput Stress Phenotyping in Plants." *Trends in Plant Science* 21 (2): 110–24.

Sladojevic, Srdjan, Marko Arsenovic, Andras Anderla, Dubravko Culibrk, and Darko Stefanovic. 2016. "Deep Neural Networks Based Recognition of Plant Diseases by Leaf Image Classification." *Computational Intelligence and Neuroscience* 2016.

Yalcin, Hulya, and Salar Razavi. 2016. "Plant Classification Using Convolutional Neural Networks." In *2016 Fifth International Conference on Agro-Geoinformatics (Agro-Geoinformatics)*, 1–5.

Zhu, Xiaojin, and Andrew B Goldberg. 2009. "Introduction to Semi-Supervised Learning." *Synthesis Lectures on Artificial Intelligence and Machine Learning* 3 (1): 1–130.